\documentclass[10pt,letterpaper]{article}
\usepackage[top=0.85in,left=2.75in,footskip=0.75in]{geometry}

\usepackage{amsmath,amssymb}
\usepackage{changepage}
\usepackage{textcomp,marvosym}
\usepackage{cite}
\usepackage{nameref,hyperref}
\usepackage[right]{lineno}
\usepackage[nopatch=eqnum]{microtype}
\DisableLigatures[f]{encoding = *, family = * }
\usepackage[table]{xcolor}
\usepackage{array}

\newcolumntype{+}{!{\vrule width 2pt}}
\newlength\savedwidth

\raggedright
\usepackage[aboveskip=1pt,labelfont=bf,labelsep=period,justification=raggedright,singlelinecheck=off]{caption}

\makeatletter
\renewcommand{\@biblabel}[1]{\quad#1.}
\makeatother

\usepackage{lastpage,fancyhdr,graphicx}
\usepackage{epstopdf}
\fancyheadoffset[L]{2.25in}
\fancyfootoffset[L]{2.25in}
\usepackage{adjustbox}
\usepackage{booktabs}
\usepackage{longtable}
\usepackage{multirow}
\usepackage{makecell}
\usepackage{enumitem}
\usepackage{rotating}
\usepackage{flafter}

\newcommand{\splittop}{\cmidrule[\heavyrulewidth](lr){1-4}\cmidrule[\heavyrulewidth](lr){5-8}\cmidrule[\heavyrulewidth](lr){9-12}\cmidrule[\heavyrulewidth](lr){13-13}}
\newcommand{\splitmid}{\cmidrule(lr){1-4}\cmidrule(lr){5-8}\cmidrule(lr){9-12}\cmidrule(lr){13-13}}
\newcommand{\splitsubmid}{\cmidrule(lr){2-4}\cmidrule(lr){5-8}\cmidrule(lr){9-12}\cmidrule(lr){13-13}}

\begin{document}
\vspace*{0.2in}

\begin{flushleft}
{\Large
\textbf{AIGaitor: Privacy-preserving and cloud-free motion analysis for everyone, using edge computing}
}
\newline
\\
Lauhitya Reddy\textsuperscript{1},
Trisha M. Kesar\textsuperscript{2,3\ddag},
Hyeokhyen Kwon\textsuperscript{1,3\ddag *}
\\
\bigskip
\textbf{1} Department of Biomedical Informatics, Emory University, Atlanta, Georgia, United States of America
\\
\textbf{2} Department of Rehabilitation Medicine, Emory University, Atlanta, Georgia, United States of America
\\
\textbf{3} The Wallace H. Coulter Department of Biomedical Engineering, Emory University and Georgia Institute of Technology, Atlanta, Georgia, United States of America
\\
\bigskip
\ddag These authors are co-senior authors.
\\
* hyeokhyen.kwon@emory.edu
\end{flushleft}

\section*{Abstract}
Motion capture is the gold standard for measuring human movements, but its clinical use remains limited by expensive laboratory-based systems and the time and technical expertise required to collect and analyze data.
Over the past decade, research has moved from marker-based to markerless motion capture, and from multi-camera arrays to a single smartphone video, with monocular systems now reporting clinically viable and accurate whole-body kinematics.
However, current monocular systems depend on dedicated GPU infrastructure, either through third-party cloud services or locally maintained workstations, which preserves old barriers of hardware cost and technical complexity while adding new limitations such as network-transfer delays, reliance on internet connectivity, and the privacy or security risks of transmitting clinical videos.
We developed \textbf{AIGaitor}, a privacy-preserving and cloud-free motion analysis system that runs all components of existing markerless motion-capture pipelines on a consumer smartphone using on-device neural accelerators.
To motivate the design of AIGaitor, we surveyed 74 rehabilitation clinicians and found that 92\% would adopt an accurate, cost-effective, easy-to-use AI gait analysis tool, with 79.7\% citing cost of operation, 68.9\% insufficient training, and 64.9\% privacy concerns as leading adoption barriers.
With these findings, we optimized and benchmarked the components of current monocular pipelines on mobile phones (iOS), spanning 2D and 3D pose estimation, pose optimization, skeleton-based deep-learning analysis, and a vision-language model.
We show that a Time-Priority end-to-end on-device pipeline processes a 10\,s 4K 60\,fps video clip in 77\,s on the iPhone 14, matching or beating the same pipeline executed on a high-end NVIDIA H200 cloud server (94\,s at the global mobile-average uplink, 66\,s at developed-world Wi-Fi).
Lightweight models such as ViTPose-s achieve real-time keypoint extraction, and skeleton-based action-recognition models deliver sub-millisecond gait classification on the same clip.
To our knowledge, AIGaitor is the first monocular system to demonstrate end-to-end on-device motion capture and downstream deep-learning analysis, a step toward clinically applicable movement analysis that is low-cost, privacy-preserving, and accessible to every individual with a smartphone.

\clearpage
\newgeometry{top=0.85in,left=1in,right=1in,footskip=0.75in}

\section*{Introduction}
\label{sec:intro}

Quantitative evaluation of movement impairments using marker-based 3D-motion capture is a gold standard measurement approach across rehabilitation~\cite{baker2006gait}, neurology~\cite{baker2006gait}, and orthopedics~\cite{kay2000gaitanalysis,wren2011efficacy}. 
Marker-based 3D-motion capture systems derive joint level kinematics by tracking anatomical landmarks using multiple cameras and reflective markers~\cite{stebbins2023clinical,colyer2018markerless}. 
However, these systems are usually restricted to research laboratories and specialized regional centers that can afford the expensive hardware as well as the technical specialists required to operate the equipment and interpret the complex output data. 
On the other hand, observational assessment of gait through visual inspection is low cost but has limited accuracy, test-retest, and inter-rater reliability, varying with clinician experience levels~\cite{krebs1985reliability,brunnekreef2005reliability,toro2003review,eastlack1991interrater,kim2011reliability}.
To address these gaps, recent research leveraged advances in deep-learning-based 2D and 3D human pose estimation algorithms~\cite{cao2019openpose,sun2019hrnet,mathis2018deeplabcut,pavllo2019videopose3d} and parametric body models~\cite{loper2015smpl,kanazawa2018hmr} to drive a shift from marker-based to markerless motion capture, and from multiple infrared cameras to as few as one smartphone camera. 
To accommodate for the poor accuracy of early pose estimation models and the partial occlusion of body parts involved in human movement, early markerless systems used arrays of synchronized cameras to reproduce marker-based gait kinematics within clinically acceptable error margins~\cite{kanko2021concurrent,nakano2020openpose,karashchuk2021anipose,needham2021accuracy,wren2023comparison,pagnon2022pose2simjoss}.
These multi-camera markerless systems lowered cost and could operate outdoors (unlike the marker-based infrared cameras), but still require expensive hardware and trained specialists to operate them.
Several recent systems further reduced the number of cameras to just two or three consumer cameras, with corresponding increases in accessibility~\cite{uhlrich2023opencap}. 
One such system, OpenCap~\cite{uhlrich2023opencap} has enabled 14,000 researchers to collect 400,000 motion trials in the three years since its release~\cite{gilon2026monocular}.
Emerging research has brought technical feasibility to single-camera markerless motion capture, validating the accuracy of monocular video against marker-based methods in deriving clinical metrics (cadence, step length, etc.) ~\cite{stenum2021twodim,drazan2021moving,stenum2024clinical} as well as whole-body kinematics~\cite{gilon2026monocular,biopose2025,peiffer2025pbl}. 
Single-view smartphone based motion capture systems brought the cost of recording to under \$1{,}000 (excluding the expense of maintaining a cloud server) and reduced the data collection requirement to a point-and-shoot smartphone video, making clinically useful motion capture possible in any setting that can record and send videos to a dedicated GPU-based server, which is a major cost driver with skyrocketing GPU prices.

Cloud processing is undoubtedly valuable, allowing researchers to analyze raw video using the increasingly larger computer vision models reaching several billion parameters in size and using similarly larger inputs (HD/4k input)~\cite{khirodkar2026sapiens2}.
But requiring cloud processing creates a hard ceiling on access and introduces a recurring cost burden.
The massive demand on GPU servers in the ongoing AI boom have resulted in even 3-year-old GPUs (H100s) experiencing rental prices spike by 40\% in the last 6 months~\cite{nishball2026gpushortage}.
To our knowledge, all current markerless clinical motion capture systems depend on dedicated GPU infrastructure, either as a third-party cloud service~\cite{uhlrich2023opencap,gilon2026monocular} or through locally configured high-performance workstations that clinics must purchase, maintain, and troubleshoot~\cite{biopose2025,peiffer2025pbl}.
As a result, today's monocular markerless motion capture systems continue to face some of the old barriers of laboratory-based motion capture, including expensive hardware and the technical complexity limiting clinical use, while also introducing new limitations such as network-transfer delays, requirement for reliable internet connectivity, and privacy and security infrastructure for transmitting identifiable videos from clinics over the network.
The privacy issues associated with transmitting videos over the cloud are tangible.
Healthcare has remained the most costly sector for data breaches, averaging \$9.77M per incident in 2024~\cite{ibm2024costofbreach}, and misconfigured cloud storage has repeatedly exposed patient images and videos~\cite{hhs2024breachportal}.
The connectivity gap is another major barrier.
As of late 2024, the International Telecommunication Union estimates that 2.6 billion people remain entirely offline~\cite{itu2024facts}, with only a small fraction of low-income countries being online. 
Average mobile-broadband traffic is eight times lower in low-income versus high-income economies, placing cloud-dependent tools out of reach for an enormous segment of the world~\cite{crooks2026lmicAI}.
Due to these limitations, cloud-dependent markerless motion capture systems serve the fraction of users who have budgets, connectivity, institutional approval, and patient consent to transmit clinical video, while the vast majority of clinical encounters, where these conditions are not met, go entirely unrecorded.

\begin{figure}[!t]
\caption{\textbf{Existing cloud-based pipeline versus AIGaitor on-device pipeline.} Existing cloud-based pipeline (top) vs.\ AIGaitor on-device pipeline (bottom) for video-based motion capture and downstream analysis. The cloud pipeline (top) requires transmitting patient video to a dedicated GPU on the cloud or on an expensive workstation, introducing privacy risks, network transfer delays, and requiring network connectivity for real-time analysis. The on-device pipeline runs the full extraction chain locally on the phone, preserving privacy by preventing video from being saved or transferred outside the device.}
\label{fig:pipeline}
\centering
\includegraphics[width=\linewidth]{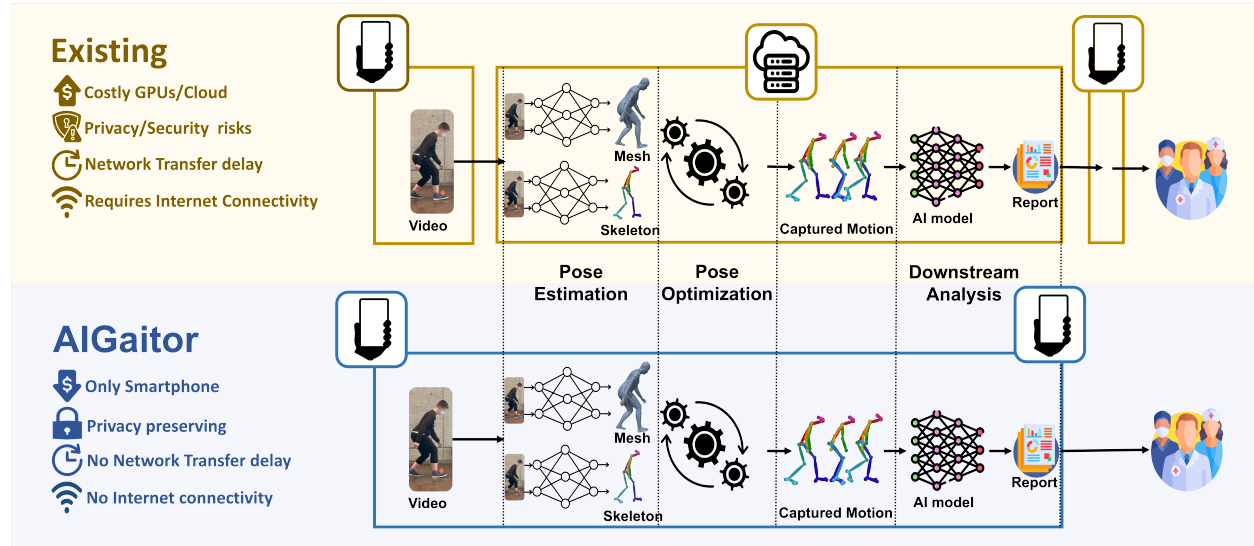}
\end{figure}
\begin{figure}[!t]
\caption{\textbf{On-device pose extraction reduces output data volume.} File size (log scale) for a 10\,s, 60\,fps, 4K clip. On-device pose extraction reduces data volume by two to three orders of magnitude vs.\ source video.}
\label{fig:datasize}
\centering
\includegraphics[width=\linewidth]{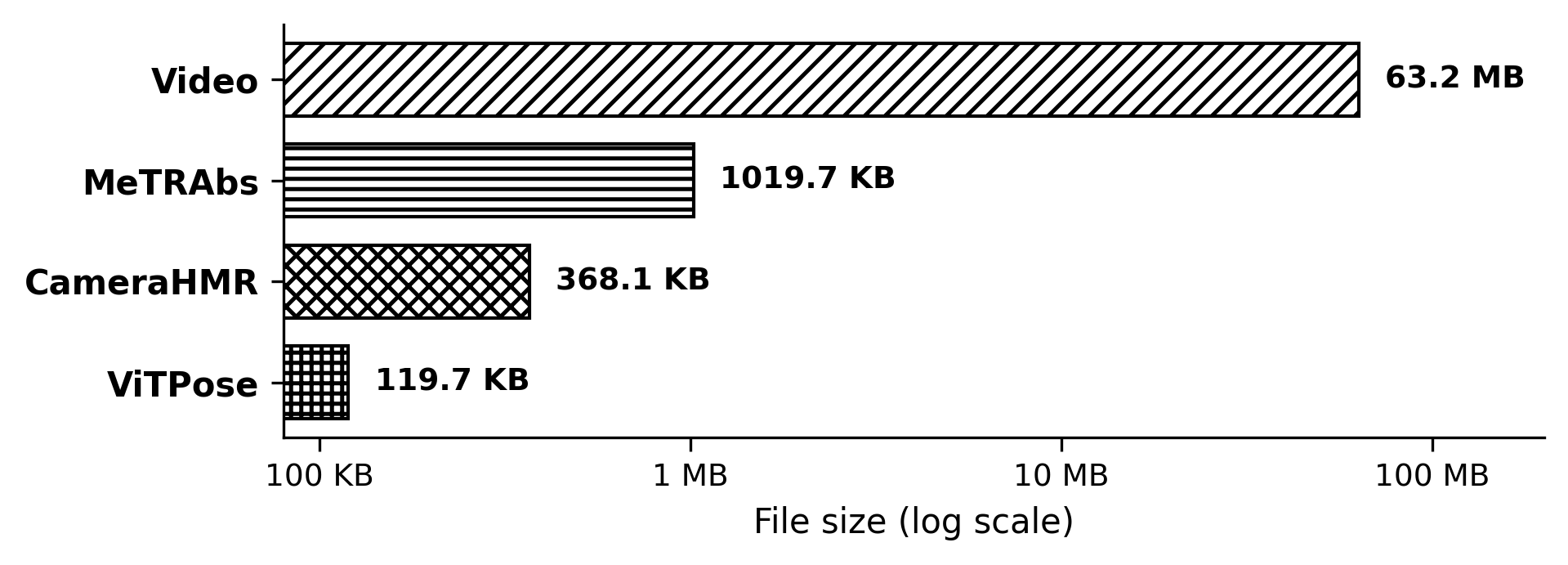}
\end{figure}

%
We propose that the natural next step in improving accessibility of video-based motion capture is to deploy these systems without dedicated external GPU servers, which avoids the recurring cost and risk of accidental video sharing on the cloud or social media~\cite{kutschera2023incidental,chawdhry2013dataprivacy,cyberhelpline_accidentalshare}.
On-device pose estimation has been a focus of mobile computer vision for several years, with multiple models demonstrated to run on consumer phones~\cite{bazarevsky2020blazepose,grishchenko2022blazeposeghum,jiang2023rtmpose}, but these lightweight models have not been adopted in clinical settings because they use limited keypoint sets insufficient for accurate full-body biomechanics, and remain less accurate than the ViTPose and HMR-family models that drive current state of the art cloud-based clinical motion capture systems.
Fortunately, the past decade has seen the maturation of mobile neural accelerators that now deliver 35--100 tera floating-point operations per second on consumer phones~\cite{apple2026ane,qualcomm2026hexagon}, alongside compilers like CoreML~\cite{apple2023coreml} and TensorFlow Lite~\cite{david2021tflite} that enable optimization of PyTorch and TensorFlow models for on-device execution.
Furthermore, the best monocular clinical motion capture pipelines today stop at generating complex kinematic data that still require substantial time and biomechanics expertise to interpret~\cite{uhlrich2023opencap,gilon2026monocular,peiffer2025pbl}. 
However, recent papers have shown that skeleton-based deep learning models can perform various clinical classification and diagnostic tasks automatically on extracted pose data~\cite{tian2022abnormalgait,zhang2023wmstgcn,jun2023hybridgait}.
Thus, we built \textbf{AIGaitor}, an iOS application (Fig~\ref{fig:pipeline}) that demonstrates the feasibility of running multiple model families used across current monocular clinical motion capture pipelines entirely on-device, including 2D pose~\cite{xu2022vitpose}, 3D pose~\cite{sarandi2020metrabs,sarandi2023acae}, mesh recovery~\cite{goel2023fourdhuman,patel2024camerahmr}, skeleton-based downstream analysis~\cite{ordonez2016deepconvlstm,shi2019twostreamagcn,dosovitskiy2021vit}, and a vision-language model in line with emerging interpretability work in systems such as BiomechGPT~\cite{yang2025biomechgpt,gemma4}.
To our knowledge, no prior work has integrated this full stack onto consumer mobile hardware, with the closest comparators either pushing pose extraction to the cloud~\cite{uhlrich2023opencap,gilon2026monocular,peiffer2025pbl} or running on-device but stopping at 2D or 3D pose without mesh recovery or downstream analysis~\cite{bazarevsky2020blazepose,xu2024mobileposer}.
By moving compute on-device, AIGaitor eliminates the recurring cost of cloud or local GPU infrastructure on a device that 5.6 billion people already carry~\cite{gsma2024mobile}, keeps raw video containing Personal Health Identifiers (PHI) on the recording device with optional real-time processing that requires no local storage, and reduces the data that leave the device to pose keypoints having two to three orders of magnitude smaller than the raw source video (Fig~\ref{fig:datasize}) and containing no pixels, identifiable features, faces, or background imagery.
The cloud is still important but with a different primary role, as on-device processing becomes available to every clinician and every patient, and cloud processing remains an opt-in for the subset of users who consent to share their video data, which we hope enables low-burden data collection at the scale of millions of hours.
By unifying upstream pose, mesh, and biomechanics extraction with downstream skeleton-based analysis and language-model interpretation on a single phone  using edge computing, AIGaitor establishes the technical foundation for a fully interpretable gait analysis tool that produces clinically actionable outputs while being privacy preserving, low-cost and accessible.

\section*{Materials and methods}
\label{sec:methods}
The current study presents a clinician survey pertinent to barriers and facilitators of AI-based gait analysis tools, and introduces our AIGaitor mobile system and its performance benchmarked against conventional cloud-based systems. 
\subsection*{Ethics statement}
All study procedures involving human participants (survey) were reviewed and approved by the Emory University Institutional Review Board (Protocol \#:
2025P009490). All participants provided informed consent prior to enrollment.
\subsection*{Clinician survey}
We administered a web-based survey via Qualtrics over a twelve-month period (February 2025 -- February 2026) to licensed physical therapy (PT) clinicians and third-year Doctor of Physical Therapy students recruited from the University and local community. 
Participants completed the online survey, which comprised four sections: (1)~participant demographics and self-rated confidence in observational gait analysis, (2 and 3)~observational gait analysis, where clinicians viewed a clinical gait video and answered multiple questions; and (4) an AI survey section assessing clinician awareness, attitudes, perceived barriers, desired outputs, and openness regarding AI-based gait video analysis tools.
The full list of AI section questions is provided in S1 Appendix.
The current paper reports on the AI section (4) and the patient demographics (1) survey sub-sections only. 

\subsection*{iOS application}
The AIGaitor application was written using Swift~\cite{swift2015}, a language for iOS native applications.
The application was built on the Xcode~\cite{apple2026xcode} development environment, before being tested and deployed as a TestFlight~\cite{apple2024testflight} application to several recent iOS device generations.

\subsection*{Model deployment to iOS}
\begin{figure}[!t]
\caption{\textbf{AIGaitor end-to-end user workflow.} AIGaitor end-to-end user workflow from video recording and trimming to Quality-Priority or Time-Priority processing and downstream analysis with joint angles and AI-generated reporting entirely on mobile phone.}
\label{fig:workflow}
\centering
\includegraphics[width=\linewidth]{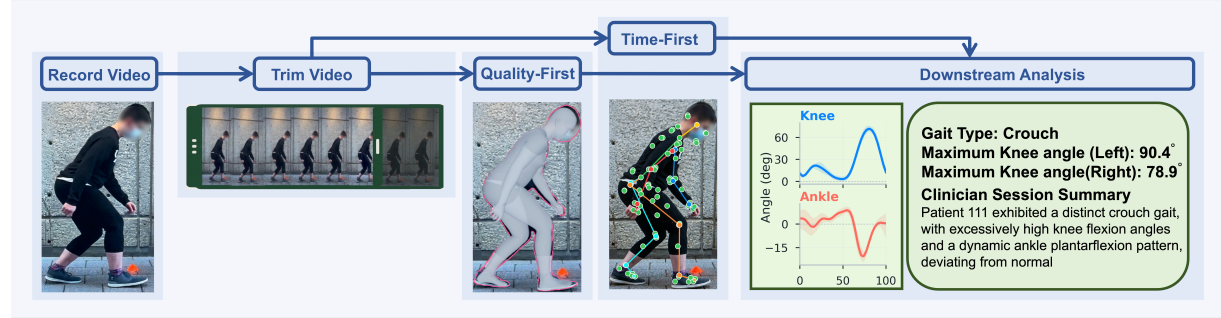}
\end{figure}

We identified common components of existing monocular motion capture systems and focused on optimizing them for iOS devices.
For 2D/3D pose estimation AIGaitor deploys ViTPose~\cite{xu2022vitpose} and MeTRAbs-ACAE~\cite{sarandi2020metrabs,sarandi2023acae}.
For HMR/SMPL mesh recovery it runs WHAM~\cite{shin2024wham}, HMR2.0~\cite{goel2023fourdhuman}, and CameraHMR~\cite{patel2024camerahmr}.
Regarding the HMR models, we use the FastHMR variants~\cite{mehraban2026fasthmr} of HMR2.0 and CameraHMR, which apply token and layer merging with a diffusion decoder to deliver up to 2.3$\times$ speedup at matching or slightly improved mesh-recovery accuracy compared to the original models.
For Downstream analysis, we deploy three example downstream skeleton-based action recognition models for gait analysis tasks like classification of gait type and regression of clinical scores: DeepConvLSTM~\cite{ordonez2016deepconvlstm}, two-stream Adaptive Graph Convolutional Networks (AGCN)~\cite{shi2019twostreamagcn}, and a hierarchical model that passes sequence of embeddings from the penultimate layer of AGCN as tokens through a Vision Transformer (ViT)~\cite{dosovitskiy2021vit}.
These families have prior validation in conventional skeleton-based activity recognition ~\cite{ordonez2016deepconvlstm, shi2019twostreamagcn}, and their clinical applications \cite{kwon2023fogagcn}, as ViTs are a powerful multi-purpose tool in Deep learning~\cite{dosovitskiy2021vit}.
Finally, for interpretation and natural-language tasks we deploy the state-of-the-art Gemma 4 Vision Language Model (VLM)~\cite{gemma4}. 
Models are deployed via CoreML~\cite{apple2023coreml} on iOS.
All neural network models were optimized from their published FP32 PyTorch checkpoints to a CoreML-compatible version for on-device iOS inference on the Apple Neural Engine (ANE).
For each model, we constructed an inference-only wrapper around the trained network, removing training-only state before optimization.
This is done by tracing the set of operations involved in inference to construct a graph of operations after which they are converted into CoreML-compatible and optimized equivalents.
As a preprocessing to inference on any of the models, we identify and pass only the bounding boxes around the humans in the scene using the Apple Vision Framework's~\cite{apple2024visionframework} \texttt{VNDetectHumanRectanglesRequest} method, which is a fast native solution running at 1-2 ms per frame.
Where practical, preprocessing and output postprocessing was also traced into the CoreML graph to reduce latency (e.g., data transfer I/O) by preventing data from moving unnecessarily between different components of the A15 Bionic chip, which has a GPU, CPU, and ANE.
The ViTPose CoreML model includes image normalization and heatmap argmax decoding, the MeTRAbs CoreML model directly returns 2D/3D coordinate tensors, and the CameraHMR CoreML model returns SMPL pose, shape, and camera parameters directly.
Outside the model, we handle all data processing in Swift~\cite{swift2015}, Apple's native language for Apple devices, primarily leveraging the Accelerate framework~\cite{apple2024accelerate} for computational speedup.
Representative outputs from each on-device pipeline stage are shown in Fig~\ref{fig:workflow}.

\subsection*{AIGaitor vs cloud pipeline benchmarking}
AIGaitor is benchmarked on an iPhone 14 (A15 Bionic~\cite{apple2021a15iphone13,apple2023iphone14specs}) as a widely used prior-generation device that provides a conservative expectation of available iOS performance, against a high-end cloud baseline running on an NVIDIA H200 NVL 143\,GB GPU~\cite{nvidia2024h200datasheet} and an Intel Xeon 6731P~\cite{intel2025xeon6731p} with 32 cores and 48\,GB of RAM allocated during inference.
All cloud models were optimized for NVIDIA GPUs by compiling with \texttt{torch.compile}~\cite{ansel2024pytorch2} and used a batch size of 64, while the on-device CoreML optimized models used a batch size of 8 and were loaded onto the device through a TestFlight~\cite{apple2024testflight} version of our application.
We compared AIGaitor and cloud end-to-end forward pass latencies and processing times for a 10-second 60\,fps gait video.
The forward pass latencies per frame (ms/frame) are an average recorded over a 10-minute 4K 60\,fps video stream of a single individual, with the long-running time accounting for any on-device thermal throttling.
We evaluate forward pass latencies of each  component of AIGaitor (pose estimation, pose optimization, downstream analysis with skeleton deep learning models, and downstream analysis with a VLM).

The Pose Estimation task loads a 10\,s 4K 60\,fps clip into memory, obtains per-frame bounding boxes via YOLO11m on the cloud and Apple Vision's \texttt{VNDetectHumanRectanglesRequest}~\cite{apple2024visionframework} on AIGaitor, crops each frame to the detected box, passes the crops through the pose model in forward passes, and stores the resulting pose data as an \texttt{.npy} file in both cloud and AIGaitor, with the latter using a swift based numpy converter.
The Pose Optimization benchmark assumes that upstream detections and initial pose estimates have already been computed, and measures only the sequence-level refinement applied to the resulting 10\,s pose track.
For WHAM, following the original implementation, we benchmark only the learned `temporal refinement' component conditioned on precomputed ViTPose-H and CameraHMR outputs \cite{shin2024wham}, whereas for `pose refinement', we benchmark the OpenCap Monocular refinement operations that jointly refine SMPL pose, translation, and shape over the full sequence using reprojection (onto 2D pose) and biomechanical consistency losses as detailed in Gilon \textit{et al} \cite{gilon2026monocular}.
WHAM is optimized as a CoreML model running on the ANE and similarly as a PyTorch model on the GPU, the `pose refinement' operations are replicated in Swift and run on CPU, while the cloud pipeline uses the functions written as a differentiable loss in PyTorch.
The Downstream Analysis deep learning model benchmark also assumes pose estimates already exist, handles preprocessing and windowing, and produces a single trial-level score by naive majority voting across windows for DeepConvLSTM and Adaptive-GCN, or directly from AGCN+ViT output for the hierarchical AGCN+ViT model, with preprocessing, dataset, and classification task identical to our previous work~\cite{reddy2025classifying}.
The Gemma 4 task takes a 500-word description of the clinical history of the patient, along with gait parameters and deep learning outputs (e.g., knee and ankle angles, gait types) and produces a 200-word plain-language summary as text output, with the table's $\downarrow$\,in and $\uparrow$\,out columns reporting the model's forward-pass speed in ms/token for input prefill and output decode respectively.
For the end-to-end benchmark, our \textit{Quality-Priority} pipeline involves mesh recovery with CameraHMR (FastHMR)~\cite{mehraban2026fasthmr}, pose refinement using ViTPose-L following the procedure described in the OpenCap monocular paper~\cite{gilon2026monocular}, AGCN classification of the refined pose output, and clinician-facing rendering of the pose overlay and classification banner.
On the other hand, the \textit{Time-Priority} pipeline replaces CameraHMR with the smaller MeTRAbs-L 3D pose model while keeping the same ViTPose-L-driven pose refinement procedure, AGCN classification of the refined pose output, and clinician-facing rendering path.
This pipeline is intended to represent a low-latency path from raw video to a rendered output, using models and methods existing in the literature \cite{gilon2026monocular,peiffer2025pbl}.
The on-device variant renders this output directly without saving the source video, while the cloud variant additionally incurs a 4K video upload before preprocessing and analysis result download to the clinician's mobile device for replay.

\section*{Results}
\label{sec:results}
\subsection*{Survey results}
Of the 74 respondents at the end of our survey, 38 were licensed physical therapists with a mean of 10.0 years clinical experience and 36 were Doctor of Physical Therapy (DPT) students from Emory University, of whom 56 were female and 18 male.
Forty respondents were actively treating patients at the time of the survey, with self-rated confidence in observational gait analysis of $3.87 \pm 0.75$ on a 5-point scale.
98.6\% of clinicians were aware of AI but only 5.4\% had prior hands-on use of these tools for gait analysis.
On a 5-point Likert scale (1 = strongly disagree, 5 = strongly agree), respondents rated the statement that AI could improve observational gait analysis accuracy at $4.09 \pm 0.72$ and that AI could detect subtle gait deviations at $3.89 \pm 0.77$, while trust in AI-generated assessments was rated $3.49 \pm 0.67$, concern about privacy and data security $3.38 \pm 0.92$, expected benefit of AI in clinical practice $3.20 \pm 0.74$, and expected benefit in research $3.27 \pm 0.69$.
In descending order of frequency, the barriers to adoption were: technology cost (79.7\%), lack of evidence or guidelines (77.0\%), insufficient training or expertise (68.9\%), patient data privacy (64.9\%), limited time or resources (44.6\%), technical complexity in usage (40.5\%), and other (13.5\%).
In descending order of frequency, the desired outputs from an AI-based gait app were: playback of the most affected phase (83.8\%), identification of the most affected gait-cycle phase (77.0\%), biomechanical information including joint angles (75.7\%), and classification of the pathological gait pattern type (73.0\%).
For overall desire to adopt, 91.9\% indicated yes, 6.8\% indicated unsure, and 1.4\% indicated no.
The full breakdown across the five survey cohorts is reported in Table~\ref{tab:ai_survey}.

\begin{table}[!t]
\centering
\caption{\textbf{Detailed AI-section results across all 5 survey cohorts ($N = 74$).} Likert items are scored from 1 for Strongly Disagree to 5 for Strongly Agree, and the ``Agree'' column reports the percentage choosing 4 or 5. Multi-select items report the count and percentage of respondents who selected each option.}
\label{tab:ai_survey}
\small
\renewcommand{\arraystretch}{1.15}
\begin{tabular}{@{} l r @{}}
\toprule
\textbf{Question (short version)} & \textbf{Result} \\
\midrule
\multicolumn{2}{@{}l}{\textit{Awareness, prior use, and adoption intent (Yes / No / Unsure)}} \\
Aware of what Artificial Intelligence is                         & 73 (98.6\%) Yes; 1 (1.4\%) No \\
Currently using AI-based tools for gait analysis                 & 4 (5.4\%) Yes; 70 (94.6\%) No \\
Would adopt AI gait tools if accurate, cost-effective, easy      & 68 (91.9\%) Yes; 5 (6.8\%) Unsure; 1 (1.4\%) No \\
\midrule
\multicolumn{2}{@{}l}{\textit{Likert (1--5): mean $\pm$ SD\ \ ;\ \ \% who agree (4 or 5)}} \\
AI can improve OGA accuracy                                       & $4.09 \pm 0.72$\ ;\ 87.8\% \\
AI can detect subtle gait deviations missed by human OGA          & $3.89 \pm 0.77$\ ;\ 79.7\% \\
Trust AI-generated gait assessments in clinical decisions         & $3.49 \pm 0.67$\ ;\ 50.0\% \\
Concern about privacy and data security in clinical AI            & $3.38 \pm 0.92$\ ;\ 48.6\% \\
Expected benefit of AI gait tools in clinical practice            & $3.20 \pm 0.74$\ ;\ 39.2\% \\
Expected benefit of AI gait tools in research                     & $3.27 \pm 0.69$\ ;\ 40.5\% \\
\midrule
\multicolumn{2}{@{}l}{\textit{Main barriers to adopting AI-based gait analysis (multi-select, $N=74$)}} \\
Cost of technology                                                & 59 (79.7\%) \\
Lack of clear evidence or guidelines                              & 57 (77.0\%) \\
Insufficient training or expertise                                & 51 (68.9\%) \\
Concern about patient data privacy                                & 48 (64.9\%) \\
Limited time or resources                                         & 33 (44.6\%) \\
Technical complexity in usage                                     & 30 (40.5\%) \\
Other                                                             & 10 (13.5\%) \\
\midrule
\multicolumn{2}{@{}l}{\textit{Minimum outputs desired from an AI-based gait app (multi-select, $N=74$)}} \\
Playback / replay of the most affected phase                      & 62 (83.8\%) \\
Which phase of the gait cycle is most affected                    & 57 (77.0\%) \\
Biomechanical information (e.g., joint angles)                    & 56 (75.7\%) \\
Type of pathological gait pattern                                 & 54 (73.0\%) \\
Other                                                             & 3 (4.1\%) \\
\bottomrule
\end{tabular}
\renewcommand{\arraystretch}{1.0}
\end{table}

\subsection*{AIGaitor performance compared to conventional cloud-based systems}
On the iPhone 14, the four ViTPose 2D pose variants (24\,M to 632\,M parameters) ran at forward-pass latencies of 8.3 to 100\,ms/frame and completed a 10\,s 4K 60\,fps clip in 16.4 to 170.8\,s, while on the cloud pipeline the same models ran at 2.65 to 7.97\,ms/frame for clip processing times of 35.7 to 46.7\,s (Table~\ref{tab:models}).
The four 3D pose and mesh recovery models (MeTRAbs-S, MeTRAbs-L, HMR2.0, and CameraHMR), with parameter counts ranging from 24\,M to 592\,M, ran at 5 to 183\,ms/frame on-device for clip processing times of 15.9 to 139.2\,s, and at 1.3 to 12.6\,ms/frame on cloud for clip processing times of 27.2 to 42.1\,s.
In the pose optimization stage, WHAM (22\,M) ran at 26.6\,ms/frame on-device and 1.22\,ms/frame on cloud for clip processing times of 15.9 and 26.1\,s respectively, and the pose refinement step took 19.8\,s on-device and 5.7\,s on cloud.
Across nine 90-frame windows of 2D keypoints extracted with a 60-frame stride from a 10\,s clip, the three skeleton-based deep-learning models (DeepConvLSTM (1\,M), Adaptive-GCN (1\,M), and the hierarchical AGCN+ViT model (44\,M)) ran in 0.005, 0.006, and 0.017\,s on-device and in 0.03, 0.03, and 0.040\,s on cloud.
For the fixed VLM summarization task (a 500-word clinical description condensed into a 200-word output, approximately 750 input and 300 output tokens), Gemma 4-E2B (2\,B) and Gemma 4-E4B (4\,B) completed the task in 29.2 and 55.7\,s on-device with input ($\downarrow$) and output ($\uparrow$) forward-pass rates of 5.7 and 83\,ms/token for E2B and 17 and 143\,ms/token for E4B (Table~\ref{tab:models}), while the corresponding cloud completion times were 2.4 and 4.8\,s at 0.33 and 7.1\,ms/token for E2B and 0.67 and 14.3\,ms/token for E4B.
The VLM rows show a striking input-output cost asymmetry, with output decoding costing 14 to 17$\times$ more per token than input encoding on-device, so on-device VLM deployment is practical for tasks with short structured outputs (the templated short-form notes that dominate physical-therapy and outpatient progress-note documentation~\cite{rule2021notes}) and impractical for long free-form or iterative generation.
This asymmetry arises because input prefill evaluates the prompt in a parallel forward pass, whereas output decoding is autoregressive and must rerun the model sequentially for each generated token.
The slower on-device input prefill relative to cloud also reflects mobile memory limits, since much of device memory is occupied by the model weights and the remaining activation budget constrains how many prompt tokens can be processed in parallel.
Uploading a single 10\,s 4K 60\,fps clip (27.7\,MB) took 16\,s at 15\,Mbps (global mobile-average uplink) and 2.0\,s at 300\,Mbps (Fast, developed-world Wi-Fi).

\begin{sidewaystable}[p]
\centering
\caption{\textbf{Computational benchmarks comparing AIGaitor (iPhone 14, A15 Bionic) with a cloud pipeline (NVIDIA H200 NVL 143\,GB GPU, Intel Xeon 6731P, 48\,GB RAM) across different model families used in monocular smartphone-based motion capture.} Rows are grouped by pipeline processing stage. Each row reports model forward pass latency (ms/frame, or ms/token for VLM with $\downarrow$\,in prefill and $\uparrow$\,out decode) and total processing time for a 10\,s 4K 60\,fps clip. $^\ddagger$DL models operate on 90-frame windows of 2D keypoints with 60-frame stride (9 windows per 10\,s clip). On-device speed up is the time taken on cloud divided by time taken on AIGaitor to process a 10\,s video.}
\label{tab:models}

\small
\setlength{\tabcolsep}{2pt}
\renewcommand{\arraystretch}{1.12}

\begin{adjustbox}{max width=\textheight, max totalheight=0.82\textwidth, center}
\begin{tabular}{@{} c l l r | c c c c | c c c c | c @{}}
\splittop
& & & & \multicolumn{4}{|c|}{\textbf{AIGaitor (iPhone 14)}} & \multicolumn{4}{c|}{\textbf{Cloud Pipeline}} & \\
\cmidrule(lr){5-8} \cmidrule(lr){9-12}
\textbf{Stage} & \textbf{Method} & \textbf{Model (Backbone)} & \textbf{Parameters}
  & \multicolumn{2}{|c}{\textbf{ms/frame}} & \multicolumn{2}{c}{\makecell{\textbf{End to End Processing time}\\\textbf{10\,s video\,(s)}}}
  & \multicolumn{2}{|c}{\textbf{ms/frame}} & \multicolumn{2}{c}{\makecell{\textbf{End to End Processing time}\\\textbf{10\,s video\,(s)}}}
  & \makecell{\textbf{AIGaitor Gain over Cloud} \\\textbf{(Cloud time/AIGaitor time)}} \\
\splitmid

\multirow{8}{*}{\rotatebox[origin=c]{90}{\scriptsize\textbf{Pose estimation}}}
  & \multirow{4}{*}{2D Pose}
    & ViTPose-S (ViT-S)         & 24M  & \multicolumn{2}{|c}{8}    & \multicolumn{2}{c}{16.4}  & \multicolumn{2}{|c}{3}  & \multicolumn{2}{c}{35.9}  & 2.19$\times$ \\
  &  & ViTPose-B (ViT-B)         & 86M  & \multicolumn{2}{|c}{12}   & \multicolumn{2}{c}{20.6}  & \multicolumn{2}{|c}{3}  & \multicolumn{2}{c}{35.7}  & 1.73$\times$ \\
  &  & ViTPose-L (ViT-L)         & 200M & \multicolumn{2}{|c}{57}   & \multicolumn{2}{c}{37.9}  & \multicolumn{2}{|c}{5}  & \multicolumn{2}{c}{43.0}  & 1.13$\times$ \\
  &  & ViTPose-H (ViT-H)         & 632M & \multicolumn{2}{|c}{100}  & \multicolumn{2}{c}{170.8} & \multicolumn{2}{|c}{8}  & \multicolumn{2}{c}{46.7}  & 0.27$\times$ \\
\splitsubmid
  & \multirow{4}{*}{3D Pose / HMR}
    & MeTRAbs (EffNetV2-S)      & 24M  & \multicolumn{2}{|c}{5}    & \multicolumn{2}{c}{15.9}  & \multicolumn{2}{|c}{4}   & \multicolumn{2}{c}{38.5}  & 2.42$\times$ \\
  &  & MeTRAbs (EffNetV2-L)      & 120M & \multicolumn{2}{|c}{32}   & \multicolumn{2}{c}{25.2}  & \multicolumn{2}{|c}{13}  & \multicolumn{2}{c}{42.1}  & 1.67$\times$ \\
  &  & HMR2.0 (ViT-H)            & 592M & \multicolumn{2}{|c}{149}  & \multicolumn{2}{c}{115.3} & \multicolumn{2}{|c}{1}   & \multicolumn{2}{c}{27.2}  & 0.24$\times$ \\
  &  & CameraHMR (ViT-H)         & 553M & \multicolumn{2}{|c}{183}  & \multicolumn{2}{c}{139.2} & \multicolumn{2}{|c}{1}   & \multicolumn{2}{c}{30.5}  & 0.22$\times$ \\

\splitmid
\multirow{2}{*}{\rotatebox[origin=c]{90}{\makecell[c]{\scriptsize\textbf{Pose}\\[-0.2ex]\scriptsize\textbf{Optimization}}}}
  & \rule[-3.0ex]{0pt}{9.5ex}Temporal refinement & WHAM (LSTM)  & 22M
  & \multicolumn{2}{|c}{27}
  & \multicolumn{2}{c}{15.9}
  & \multicolumn{2}{|c}{1.22}
  & \multicolumn{2}{c}{26.1}
  & 1.64$\times$ \\

  & \rule[-3.0ex]{0pt}{9.5ex}Pose refinement & NA & NA
  & \multicolumn{2}{|c}{NA}
  & \multicolumn{2}{c}{19.8}
  & \multicolumn{2}{|c}{NA}
  & \multicolumn{2}{c}{5.7}
  & 0.29$\times$ \\

\splitmid
\multirow{7}{*}{\rotatebox[origin=c]{90}{\parbox{2.6cm}{\centering\scriptsize\textbf{Downstream analysis}}}}
  & \multirow{3}{*}{DL Models$^\ddagger$}
    & DeepConvLSTM              & 1M   & \multicolumn{2}{|c}{0.005}  & \multicolumn{2}{c}{0.005} & \multicolumn{2}{|c}{0.0071} & \multicolumn{2}{c}{0.03}  & 6.00$\times$ \\
  &  & Adaptive-GCN               & 1M   & \multicolumn{2}{|c}{0.0067} & \multicolumn{2}{c}{0.006} & \multicolumn{2}{|c}{0.012}  & \multicolumn{2}{c}{0.03}  & 5.00$\times$ \\
  &  & Hierarchical (AGCN + ViT)  & 44M  & \multicolumn{2}{|c}{0.025}  & \multicolumn{2}{c}{0.017} & \multicolumn{2}{|c}{0.028}  & \multicolumn{2}{c}{0.040} & 2.35$\times$ \\

  & \textbf{} & \textbf{} & \textbf{}
    & \multicolumn{1}{|c}{\textbf{$\downarrow$\,in}} & \textbf{$\uparrow$\,out} & \multicolumn{2}{c}{}
    & \multicolumn{1}{|c}{\textbf{$\downarrow$\,in}} & \textbf{$\uparrow$\,out} & \multicolumn{2}{c}{}
    & \\
\cmidrule(lr){5-6} \cmidrule(lr){9-10}

  & \multirow{2}{*}{VLM}
    & Gemma 4-E2B               & 2B   & 5.7 & 83  & \multicolumn{2}{c}{29.2} & 0.33 & 7.1  & \multicolumn{2}{c}{2.4} & 0.08$\times$ \\
  &  & Gemma 4-E4B               & 4B   & 17  & 143 & \multicolumn{2}{c}{55.7} & 0.67 & 14.3 & \multicolumn{2}{c}{4.8} & 0.09$\times$ \\

\splitmid
\multirow{2}{*}{\rotatebox[origin=c]{90}{\makecell[c]{\scriptsize\textbf{Cloud}\\[-0.2ex]\scriptsize\textbf{transfer}}}}
  & \multicolumn{2}{l}{\rule{0pt}{4.8ex}4K 60\,fps video (27.7\,MB) @ 15\,Mbps (global avg)}
  &  & \multicolumn{4}{|c|}{0} & \multicolumn{4}{|c|}{16\,s} & --- \\
  & \multicolumn{2}{l}{\rule{0pt}{4.8ex}4K 60\,fps video (27.7\,MB) @ 300\,Mbps (Fast)}
  &  & \multicolumn{4}{|c|}{0} & \multicolumn{4}{|c|}{2.0\,s} & --- \\

\splitmid
\splitmid
& \multicolumn{3}{@{}l}{\rule{0pt}{4.8ex}\textbf{\textit{Time-Priority} pipeline}}
  & \multicolumn{4}{|c|}{77\,s}
  & \multicolumn{2}{|c}{\textit{Global avg}\ 94\,s}
  & \multicolumn{2}{c}{\textit{Fast}\ 66\,s}
  & 1.22$\times$ / 0.86$\times$ \\
& \multicolumn{3}{@{}l}{\scriptsize MeTRAbs-L $\to$ Pose Refinement (with ViTPose-L) $\to$ AGCN $\to$ Render}
  & & & & & & & & & \\[1ex]

& \multicolumn{3}{@{}l}{\rule{0pt}{4.8ex}\textbf{\textit{Quality-Priority} pipeline}}
  & \multicolumn{4}{|c|}{153\,s}
  & \multicolumn{2}{|c}{\textit{Global avg}\ 84\,s}
  & \multicolumn{2}{c}{\textit{Fast}\ 55\,s}
  & 0.55$\times$ / 0.36$\times$ \\
& \multicolumn{3}{@{}l}{\scriptsize CameraHMR $\to$ Pose Refinement (with ViTPose-L) $\to$ AGCN $\to$ Render}
  & & & & & & & & & \\[1ex]

\splitmid
\end{tabular}
\end{adjustbox}
\end{sidewaystable}

The \textit{Time-Priority} end-to-end pipeline (MeTRAbs-L for 3D pose followed by pose refinement using outputs from ViTPose-L, AGCN classification of the refined pose output, and rendering output onto a video replay) completed in 77\,s on the iPhone 14, 94\,s on cloud at the global mobile-average condition, and 66\,s on cloud at the Fast condition, for Cloud / AIGaitor processing-time ratios of 1.22$\times$ and 0.86$\times$ respectively.
The \textit{Quality-Priority} end-to-end pipeline (CameraHMR followed by pose refinement using outputs from ViTPose-L, AGCN classification of the refined pose output, and rendering output onto a video replay) completed in 152.6\,s on the iPhone 14, 84\,s on cloud at the global mobile-average condition, and 55\,s on cloud at the Fast condition, for Cloud / AIGaitor processing-time ratios of 0.55$\times$ and 0.36$\times$ respectively.
Across all per-stage rows the Cloud / AIGaitor processing-time ratios ranged from 0.08$\times$ (Gemma 4-E2B) to 6.00$\times$ (DeepConvLSTM), with values below 1$\times$ on ViTPose-H, both HMR variants (HMR2.0 and CameraHMR), the pose refinement step, and both Gemma 4 variants, and values above 1$\times$ on the three smaller ViTPose variants, both MeTRAbs variants, WHAM, and the three downstream DL models (Table~\ref{tab:models}, rightmost column).
End-to-end results should be interpreted separately from forward-pass latency which in everycase but DL models is faster on cloud.
For example, MeTRAbs-L had a faster cloud forward pass than the iPhone (13 vs 32\,ms/frame) but a slower cloud end-to-end processing time (42.1 vs 25.2\,s), and the same practical advantage for AIGaitor appeared across ViTPose-S, ViTPose-B, ViTPose-L, WHAM, and all three downstream DL models where fixed preprocessing and data movement costs outweighed raw model throughput.

\section*{Discussion}
\label{sec:discussion}
AIGaitor demonstrates that clinically meaningful gait-analysis pipelines can be performed fully on-device with practical latency while directly addressing key barriers limiting AI adoption in rehabilitation settings. 
Our survey findings show that clinicians recognize AI's potential to improve observational gait-analysis accuracy and detect subtle gait deviations, yet real-world use remains rare, with clinicians indicating cost, limited training, and patient-video privacy concerns (Table~\ref{tab:ai_survey}) as being responsible. 
AIGaitor's on-device design directly mitigates these barriers by avoiding recurring cloud-compute costs and eliminating the need to transmit identifiable gait video. 
Computational benchmarks further show that consumer smartphone hardware can execute end-to-end monocular motion-capture workflows within clinically feasible timeframes, with the \textit{Time-Priority} pipeline matching or exceeding cloud performance under realistic mobile-upload conditions and the \textit{Quality-Priority} pipeline remaining within a practical processing window (Table~\ref{tab:models}).
Together, these results suggest that modern mobile hardware can support scalable, privacy-preserving gait analysis without requiring cloud infrastructure for routine clinical use.

\subsection*{Survey findings}
Despite common awareness of clinical AI (98.6\%), only 5.4\% of our cohort had ever used an AI tool for gait analysis, a gap that mirrors the low early-adoption pattern documented across other clinical specialties such as the American Medical Association's 2024 physician survey of $N \approx 1{,}200$, where general AI use reached 66\%, but specialty-specific clinical tools lag well behind~\cite{ama2024aisurvey}.
Clinicians strongly endorsed AI's potential to improve observational gait analysis accuracy ($4.09 \pm 0.72$) and to detect subtle gait deviations they might miss ($3.89 \pm 0.77$), yet rated the expected benefit of AI gait tools in actual clinical practice and research at only $3.20 \pm 0.74$ and $3.27 \pm 0.69$ respectively.
We interpret this gap in potential vs. expectation for AI tools as being related to the finding that less than 6\% of clinicians were currently using AI in their clinical practice, and they are unaware of how AI tools are likely to be used (and help) in their clinical settings.
As to why AI tools are not more predominant in clinical settings, our survey responses reveal the main barriers to adoption highlighted by  clinicians (Table~\ref{tab:ai_survey}).
Notably, cost (79.7\%) and insufficient training or expertise (68.9\%) are two of the four most-frequent barriers.
Interestingly, 64.9\% of respondents flagged patient data privacy as a barrier, which is notably higher than the $\sim$31.6\% reported in similar surveys conducted among radiologists~\cite{zanardo2024euroaim}, which we attribute to the inherently identifiable nature of clinical gait video.
AIGaitor's on-device design directly addresses two barriers from this survey, eliminating the recurring per-session cost of cloud GPU infrastructure that drives the cost concern and removing the need to transmit identifiable patient video that drives the privacy concern.

\subsection*{Computational benchmarks}
Every stage of AIGaitor ran on-device in under 3 minutes, with a \textit{Quality-Priority} pipeline using ViTPose-L and CameraHMR pose extraction followed by pose refinement, AGCN classification of the refined pose output, and on-device overlay rendering processed in 152.6\,s on the iPhone 14.
The corresponding Cloud / AIGaitor ratios on this realistic pipeline were 0.55$\times$ at the global mobile-average uplink and 0.36$\times$ at developed-world Wi-Fi, showing the modest cloud advantage considering the larger server compute budget and must be weighed against H200 operating cost, network and physical-distance latency, and the fact that the system is unavailable without a functioning server.
A pragmatic \textit{Time-Priority} end-to-end pipeline that replaces the ViT-H mesh-recovery model for a lighter MeTRAbs-L 3D-pose backbone, while keeping the same ViTPose-L-driven pose refinement followed by AGCN classification of the refined pose output and on-device replay rendering, completed in 77\,s on the iPhone 14 against 94\,s on cloud at the global mobile-average uplink and 66\,s on cloud at developed-world Wi-Fi, for Cloud / AIGaitor ratios of 1.22$\times$ and 0.86$\times$ respectively.
On this realistic clinical motion-capture stack the on-device run therefore outpaces cloud at the global mobile-average uplink and trails by only 11\,s at the fastest tested condition, with the cloud advantage of the \textit{Quality-Priority} pipeline reduced once the mesh-recovery backbone is replaced by a lighter 3D-pose model.
This result should not be read as the iPhone having greater raw inference throughput, since the cloud forward passes are faster for the large neural models and for MeTRAbs-L.
Rather, the \textit{Time-Priority} stack is a sequence of smaller operations where the cloud pays repeated fixed costs for CPU video decoding, crop construction, CPU to GPU tensor transfer, kernel synchronization, and computational serialization, while the iPhone keeps CPU, GPU, ANE, and shared memory within one integrated chip reducing the costs of repeated data storage I/O transfers that the H200 discrete server GPU must perform over wired interconnects to the Intel Xeon CPU (e.g., transferring video data between RAM and VRAM for GPU processing).
Importantly, from a clinical experience, this is comparable to existing systems, with OpenCap Monocular reportedly taking 120\,s to process a 10\,s video (including kinetics) and Portable Biomechanics Laboratory taking ${\sim}66$\,s on server-grade hardware~\cite{gilon2026monocular,peiffer2025pbl}.
Downstream skeleton-based DL inference contributed negligibly to total latency (0.005 to 0.017\,s on-device per 10\,s clip) because the models are small (1 M to 44 M parameters) and operate on windowed 2D-keypoint embeddings requiring just 9 forward passes for an entire 10\,s video, an architectural choice partly driven by the current absence of large-scale labeled pose datasets for clinically meaningful populations that would justify training larger downstream models.
Within the cloud pipeline itself, both MeTRAbs variants (4 and 12.6\,ms/frame for the 24\,M and 120\,M sizes) run substantially slower per frame than the much larger ViT-H mesh models HMR2.0 and CameraHMR (1.3 and 1.4\,ms/frame at $\sim$550\,M parameters), which we attribute to MeTRAbs's older EfficientNetV2 backbone being suboptimal for current PyTorch graph compilers to optimize, since compilation gains are non-deterministic and skew toward more commonly used transformer architectures that compiler target first~\cite{ansel2024pytorch2}.
At the global mobile-average uplink of 15\,Mbps, video transfer alone contributes 16\,s of upload (with a comparable amount for downloading results back), representing 30 to 45\% of cloud end-to-end time across the pose-estimation time and an overhead that on-device processing eliminates entirely.
Taken together, these benchmarks show that the AIGaitor on-device pipeline operates with a delay that may be suitable for clinical workflows, with the cloud serving as an optional accelerator for the few largest models rather than a requirement for processing every gait video encountered.

\subsection*{Limitations and future work}\label{subsec:futurework}

This paper presents with several limitations that  will be addressed in ongoing and future work.
Our latency benchmarks (Table~\ref{tab:models}) do not include a comprehensive evaluation and implementation of quantization~\cite{jacob2018quantization,han2016deepcompression,frantar2023gptq,lin2024awq}, distillation~\cite{hinton2015distillation,sanh2019distilbert}, or model/input compression methods ~\cite{bolya2023tome,men2024shortgpt,gromov2024deeperlayers}, which means the reported on-device pipeline times, in particular those running large models as in GPU pipeline, represent a worst-case upper bound for the AIGaitor pipeline speed. 
There may be substantial improvement in computational time with the above-mentioned model compression techniques, but such efficiency gains will require rigorous evaluation of kinematic accuracy against marker-based ground truth on the trade-off between model accuracy and computational speed.
The second-ranked barrier, lack of clear evidence or guidelines (77.0\%), points to the relative recency of the field and the need for further study on usability and feasibility of the use of AIGaitor at scale.
Our ongoing work plans to co-design the user interface of AIGaitor with a subset of the 74 clinician respondents from our survey to collect their inputs and to maximize its usability and acceptability~\cite{waddell2024humancentred,ford2022doublediamond}.
The AIGaitor models were deployed and benchmarked on iOS, which holds only ${\sim}33\%$ of the global mobile-OS market and falls below ${\sim}18\%$ in Asia-Pacific, ${\sim}15\%$ in Africa, and ${\sim}9\%$ in Latin America, where Android dominates~\cite{statcounter2026mobileos}.
Future work will involve developing an android version of AIGaitor.
Next, on-device deployment is also not maintenance-free, since the AIGaitor application itself requires ongoing updates to track changes in operating systems, model runtimes, and the heterogeneous device landscape it must run on, although much of that burden is absorbed by the large externally maintained libraries (CoreML, Accelerate, Apple Vision Framework) on which the pipeline is built.
Finally, clinical tools are not just designed for the convenience of clinicians but also the specific patient populations that they serve, downstream analysis requires a convenient and low-cost way to customize model outputs for the gait impairments conditions seen by the clinics.
Therefore, AIGaitor needs a mechanism by which clinics can self-label and build models tailored to their clinical population problem, easily and at low cost.

\section*{Conclusion}
AIGaitor demonstrates that the full markerless motion-capture and analysis pipeline can run end-to-end on a consumer smartphone without dedicated cloud infrastructure.
A survey of $N=74$ physical-therapy clinicians (Table~\ref{tab:ai_survey}) confirmed that the dominant barriers to adopting AI gait tools are cost (79.7\%), lack of evidence or guidelines (77.0\%), insufficient training (68.9\%), and patient data privacy (64.9\%), two of which on-device deployment directly addresses.
Benchmarks on an iPhone 14 (A15 Bionic) show the end-to-end pipeline can process 10s gait videos within 3 minutes.
The \textit{Time-Priority} end-to-end processing from raw video to clinician-facing overlay and classification completes in 77\,s compared against 94\,s on an NVIDIA H200 cloud baseline at a 15\,Mbps global mobile-average uplink.
On the whole, AIGaitor effectively overcomes the rate-limiting constraint on cloud-based clinical markerless motion capture, demonstrating its technical feasibility for routine, privacy-preserving use at population scale everywhere for everyone.

\section*{Acknowledgments}

The authors thank the clinicians and Doctor of Physical Therapy students who participated in the survey.
Hyeokhyen Kwon is partially funded by the National Institute on Deafness and Other Communication Disorders (1R21DC021029),  Georgia CTSA Pilot Grants Program, and Shriners Children's.
Funding, infrastructure, and support provided by Center for Physical Therapy and Movement Science (CPTMS), Emory University. We would like to thank Dr. George Fulk for support and overall project assistance.

\bibliography{references}

\section*{Supporting information}

\paragraph*{S1 Appendix.}
\label{sec:supplement}
\textbf{AI survey questions.}
The following questions comprised the optional AI section (section 4) of the clinician survey. All questions were optional.

\begin{enumerate}[leftmargin=*]
\item Are you aware of what Artificial Intelligence (AI) is? (Yes / No)
\item Have you used or are you currently using any AI-based technology or software for clinical gait analysis or other assessments? (Yes / No; if yes, specify)
\item \textbf{Perceived Benefits and Concerns} (1 = Strongly Disagree, 5 = Strongly Agree):
\begin{enumerate}
    \item AI-based tools have the potential to improve accuracy of observational gait analysis.
    \item I would trust AI-generated gait assessments when forming my clinical decisions.
    \item I am concerned about privacy and data security when using AI in clinical practice.
    \item I believe AI can help detect subtle gait deviations that might be missed during standard observational gait analysis.
\end{enumerate}
\item How beneficial do you think AI-based gait analysis tools could be in clinical practice? (1 = Not beneficial at all, 5 = Extremely beneficial)
\item How beneficial do you think AI-based gait analysis tools could be in research? (1--5 scale)
\item What do you perceive as the main barriers to adopting AI-based gait analysis in your clinical setting? (Select all that apply): Cost of technology; Insufficient training or expertise; Lack of clear evidence or guidelines; Technical complexity in usage; Concern about patient data privacy; Limited time or resources; Other.
\item Which of the following information would you want at minimum from an AI-based gait assessment app? (Select all that apply): Biomechanics (e.g., joint angles); Type of pathological gait pattern; Which phase of the gait cycle is most affected; Playback/replay of the most affected phase; None; Other.
\item Would you consider adopting AI-based gait analysis tools if they were proven accurate, cost-effective, and user-friendly? (Yes / No / Unsure)
\end{enumerate}

\end{document}